\documentclass[10pt,twocolumn,letterpaper]{article}

\usepackage[pagenumbers]{cvpr} 

\usepackage{graphicx}
\usepackage{amsmath}
\usepackage{amssymb}
\usepackage{booktabs}
\usepackage{bbding}
\usepackage{multirow}
\usepackage[american]{babel}

\newcommand{\squishlist}{
 \begin{list}{$\bullet$}
  { \setlength{\itemsep}{0pt}
     \setlength{\parsep}{1pt}
     \setlength{\topsep}{1pt}
     \setlength{\partopsep}{0pt}
     \setlength{\leftmargin}{1em}
     \setlength{\labelwidth}{1em}
     \setlength{\labelsep}{0.5em} } }

\newcommand{\squishend}{
  \end{list}  }

\usepackage[pagebackref,breaklinks,colorlinks]{hyperref}

\def\cvprPaperID{2372} 
\def\confName{CVPR}
\def\confYear{2022}

\begin{document}

\title{A Unified Pruning Framework for Vision Transformers}

\author{Hao Yu \qquad Jianxin Wu\thanks{J. Wu is the corresponding author.}\\
State Key Laboratory for Novel Software Technology\\Nanjing University, Nanjing, China\\
	{\tt\small yuh@lamda.nju.edu.cn, wujx2001@gmail.com}
}

\maketitle

\begin{abstract}

Recently, vision transformer (ViT) and its variants have achieved promising performances in various computer vision tasks. Yet the high computational costs and training data requirements of ViTs limit their application in resource-constrained settings. Model compression is an effective method to speed up deep learning models, but the research of compressing ViTs has been less explored. Many previous works concentrate on reducing the number of tokens. However, this line of attack breaks down the spatial structure of ViTs and is hard to be generalized into downstream tasks. In this paper, we design a unified framework for structural pruning of both ViTs and its variants, namely UP-ViTs. Our method focuses on pruning all ViTs components while maintaining the consistency of the model structure.  Abundant experimental results show that our method can achieve high accuracy on compressed ViTs and variants, e.g., UP-DeiT-T achieves 75.79\% accuracy on ImageNet, which outperforms the vanilla DeiT-T by 3.59\% with the same computational cost. UP-PVTv2-B0 improves the accuracy of PVTv2-B0 by 4.83\% for ImageNet classification. Meanwhile, UP-ViTs maintains the consistency of the token representation and gains consistent improvements on object detection tasks.
\end{abstract}

\section{Introduction}

The transformer~\cite{Vaswani2017Attention} architecture has been widely used for natural language processing (NLP) tasks over the past years. Inspired by its excellent performance in NLP, ViT~\cite{dosovitskiy2020vit} employs a pure transformer structure to image classification tasks and achieves remarkable accuracy. Following the successes of ViT, transformer-based models have established many new state-of-the-art records in various computer vision (CV) tasks, such as image classification~\cite{Touvron2021deit}, object detection~\cite{carion2020end} and video segmentation~\cite{wang2021end}. In spite of these advances, most of these transformer-based structures suffer from large model size, huge run-time memory consumption and high computational costs. Therefore, there exists impending needs to develop and deploy lightweight and efficient vision transformers.

Network pruning is a useful technique to strike a good balance between model accuracy and inference speed and memory usage. Although numerous effective pruning algorithms have been proposed to compress and accelerate convolutional neural networks (CNNs) in CV tasks, how to prune vision transformers is a much less explored topic. Some previous NLP research efforts~\cite{katharopoulos2020transformers,guo2019star} focus on exploring new multi-head self-attention (MHSA) mechanisms or knowledge distillation algorithms. However, the most time-consuming module in a transformer is in fact the feed-forward network (FFN)~\cite{marin2021token}, but efforts in pruning FFN remain scarce. Recent ViT pruning methods~\cite{tang2021patch,marin2021token} are mainly attracted to recursively sampling informative tokens (or equivalently, their corresponding image patches) to increase the inference speed in image classification, which achieves similar accuracy as that of using all tokens with fewer computations. Unfortunately, these token sampling methods often impede the vision transformers' generalization ability on downstream tasks, e.g., object detection and instance segmentation. In addition, this kind of sampling is also difficult to be applied to NLP tasks, which limits the application domain of these methods. 

Hence, we believe that in order to successfully accelerate and slim vision transformers, we need a \emph{unified} approach which simultaneously prunes \emph{all} components in a transformer structure, does \emph{not} alter the transformer structure, generalizes well to downstream tasks with \emph{high} accuracies, and applies to not only ViTs but also its many \emph{variants}.

To fulfill these goals, we propose UP-ViTs, a unified pruning framework for vision transformers, which prunes the channels in ViTs in a unified manner, including those both inside and outside of the residual connections in all the blocks, MHSAs, FFNs, normalization layers, \emph{and} convolution layers in ViT variants. We first devise an efficient evaluation module to estimate the importance score of each filter in a pre-trained ViT model. Then based on the compression goals, all redundant channels will be simultaneously removed, which leads to a thinner structure. In particular, when compressing the attention layers, we investigate the influence of MHSA and propose a novel method for throwing channels away. We also design an effective progressive block pruning method, which removes the least important block and proposes new \emph{hybrid} blocks in ViTs. Experiments on ImageNet show that UP-ViTs significantly outperforms previous ViTs with the same or even higher throughput. Our contributions are:
\squishlist
    \item We propose a novel framework for structured compression of ViTs and its variants. Our method can be easily applied to both vanilla ViTs and its variants, and the resulting compressed models achieve higher accuracy than previous state-of-the-art ViTs and existing pruning algorithms.
    \item Our method maintains the consistency of the token representation, therefore we can generalize the compressed model to various downstream tasks. For example, we compare the accuracy of our pruned UP-PVTv2 and previous PVTv2 models in object detection, and achieve higher detection accuracy. 
\squishend


\section{Related Works}

We first briefly review the related works.

\subsection{Vision transformers}

Transformer~\cite{Vaswani2017Attention} utilizes the MHSA mechanism to build long-range dependencies between pairs of input tokens for NLP tasks, and recently it has been introduced into many computer vision tasks. ViT first \cite{dosovitskiy2020vit} shows that a standard transformer can achieve state-of-the-art accuracies in image classification tasks when the training data is sufficient (e.g., using JFT-300M). DeiT~\cite{Touvron2021deit} further proposes a token-based distillation method for training and explores existing data augmentation and regularization strategies, such as Rand-Augment~\cite{cubuk2020randaugment} and stochastic depth~\cite{Huang2016deep}. With the same architecture as ViT, DeiT is also effective using the smaller ImageNet-1K dataset. Since then many works attempt to introduce the local dependency into vision transformers. T2T-ViT~\cite{Yuan2021Tokens} introduces the tokens-to-token (T2T) module to aggregate neighboring tokens, which can be generalized into other efficient transformers like Performer~\cite{choromanski2021rethinking}. For adapting ViTs to dense prediction tasks such as object detection, Wang et al. proposed Pyramid Vision Transformer (PVT)~\cite{wang2021pyramid}, which introduces convolution and the pyramid hierarchical structure to the design of transformer backbones. They also updated their model later as PVTv2~\cite{pvtv2}, which achieves higher accuracy with less computation burden. Swin Transformer~\cite{liu2021Swin} proposes a shifted windowing scheme to compute representation. LV-ViT~\cite{jiang2021all} proposes a token labeling training method to improve the performance of ViT. In this paper, we will show that our compression method can deal with both the original ViTs and its variants.
\subsection{Compressing transformers and sparse tokens}

Training ViTs is time-consuming and the trained models are often massive, which necessitates compression algorithms. Starting in NLP applications, LayerDrop~\cite{fan2019reducing} randomly drops layers at training time, and at test time it selects sub-network to any desired depth. Behnke et al.~\cite{Behnke2020losing} iterated between pruning attention heads and fine-tuning. Li et al.~\cite{li2020train} showed that the most compute-efficient training scheme is to stop training after a small number of iterations, and then heavily compress them. Another related field is to reduce the complexity of the MHSA module via various approximations~\cite{guo2019star,katharopoulos2020transformers}.

Lately, some pruning algorithms have been proposed for ViTs. VTP~\cite{zhu2021visual} reduces embedding dimensionality by introducing control coefficients. Moreover, sparse training of ViTs seeks to adaptively identify highly informative sparse tokens. Tang et al.~\cite{tang2021patch} designed a patch slimming method to discard useless tokens. Evo-ViT~\cite{xu2021evo} updates the selected informative and uninformative tokens with different computation paths. However, these token sampling methods are unstructured and are difficult to use in downstream tasks (e.g., object detection). Jia et al.~\cite{jia2021efficient} explored a fine-grained manifold distillation approach that calculates training loss from three perspectives. NViT~\cite{yang2021nvit} observes the dimension trend of every component in ViTs. Our method is more general and may be potentially combined with them. 
\subsection{Pruning CNN channels}

Channel pruning methods remove less important channels in CNNs, and then the pruned sub-model can be fine-tuned to recover its accuracy. Hence it is important to distinguish important and redundant channels.  Li et al.~\cite{li2016pruning} pruned filters using their $\ell_{1}$-norm values. He et al.~\cite{he2017channel} proposed a LASSO based channel selection method. Luo et al.~\cite{luo2017thinet} pruned filters based on statistics computed from its next layer. After that, they further pruned the residual connection to get a wallet-shape model, which is a better model structure when the shortcut connection exists~\cite{luo2020neural}. If some filters in one layer are pruned, the output features of this layer and the input of the next layer should be changed correspondingly. He et al.~\cite{he2019filter} pruned channels based on the geometric median. In this work, we apply KL-divergence to evaluate the importance of filters in vision transformers, which has been proved to be effective in CNN~\cite{luo2020neural}. 

\section{The UP-ViTs Method}

We will now propose our UP-ViTs. First we review the standard ViT briefly in Section~\ref{sec:Architecture}, then our method to evaluate the importance of each filter in Section~\ref{sec:Calculate}. We elaborate on the details of pruning into a sub-model and fine-tuning it in Section~\ref{sec:Fine-tuning}. Finally, in Section~\ref{sec:Pruning_Blocks} we further propose an efficient algorithm to prune blocks in ViTs. 

\subsection{Architecture of ViT}  \label{sec:Architecture}

The standard transformer receives an input as a 1D sequence of token embeddings. Following this design, ViT reshapes an image into flattened sequential patches and linearly projects each patch into an embedding vector. It also concatenates the patch embeddings with a trainable classification token and adds a position embedding. Consider a ViT model with $L$ blocks. Given an image $\mathbf{x} \in \mathbb{R}^{C\times H\times W}$, ViT first reshapes it into flattened 2D patches $\mathbf{x}_p \in \mathbb{R}^{N\times(CP^2)}$, where $H\times W$ is the original input resolution, $C$ is number of input channels, $P \times P$ is the resolution of each patch, $N= HW/P^{2}$ is the number of patches. Then, the sequence of embeddings $\mathbf{x}_0$ is formed by
\begin{align}
    \mathbf{x}_{patch} &= \mathrm{FC}_{patch}(\mathbf{x}_p) \,,\\
   \mathbf{x}_0 &= [\mathbf{x}_{cls}\, |\, \mathbf{x}_{patch}] + \mathbf{x}_{pos} \,,
\end{align}
where $\mathrm{FC}_{patch}$ is the linear projection with $D$ output channels. $\mathbf{x}_{cls}\in \mathbb{R}^{1\times D}$ and $\mathbf{x}_{patch}\in \mathbb{R}^{(N + 1)\times D}$ are the learnable classification token and position embeddings, respectively, $[\cdot\, |\,\cdot]$ is the column-wise concatenation.

The embedding $\mathbf{x}_0$ is subsequently fed into the transformer blocks. ViT stacks several blocks to construct the whole model, and each consists of two parts: the attention layer and FFN. In the attention layer, ViT applies LayerNorm and three linear projections ($\mathrm{FC}_{q}$, $\mathrm{FC}_{k}$ and $\mathrm{FC}_{v}$) in parallel to generate queries $\mathbf{Q}$, keys $\mathbf{K}$ and values $\mathbf{V}$, i.e., 
\begin{align}
    \mathbf{x}_{k-1}^{\prime} &= \mathrm{LN}(\mathbf{x}_{k-1}) \,,\\
    \mathbf{Q} &= \mathrm{FC}_q(\mathbf{x}_{k-1}^{\prime}) \,, \\
    \mathbf{K} &= \mathrm{FC}_k(\mathbf{x}_{k-1}^{\prime}) \,, \\
    \mathbf{V} &= \mathrm{FC}_v(\mathbf{x}_{k-1}^{\prime}) \,,
\end{align}
where $\mathrm{LN}$ is layer normalization, $\mathbf{x}_{k-1}$ is the input embedding for $k = 1,\dots, L$.  Then, a multi-head self-attention (MHSA) and a linear projection $\mathrm{FC}_{proj}$ are performed:
\begin{equation}
  \mathbf{x}_{k-1}^{\prime\prime} = \mathbf{x}_{k-1} + \mathrm{FC}_{proj}(\mathrm{MHSA}(\mathbf{Q}, \mathbf{K}, \mathbf{V} )),
\end{equation}
where $\mathbf{x}_{k-1}^{\prime\prime}$ is the output of the attention layer in the $k$th block. The input and output dimensions of these linear projections are all $D$. Note that the MHSA module does \emph{not} contain any parameters. It reshapes the queries, keys and values and applies scaled dot-product attention (cf.~\cite{dosovitskiy2020vit} for details). Then, $\mathbf{x}_{k-1}^{\prime\prime} $ will be sent into FFN, which contains two FC layers ($\mathrm{FC}_1$ and $\mathrm{FC}_2$) with a GELU non-linearity:
\begin{align}
  \mathbf{x}_{k} &= \mathbf{x}_{k-1}^{\prime\prime} + \mathrm{MLP}(\mathrm{LN}(\mathbf{x}_{k-1}^{\prime\prime}))\,,\\
  \mathrm{MLP}(\mathbf{x}) &= \mathrm{FC}_{2}\left(\mathrm{GELU}(\mathrm{FC}_{1}(\mathbf{x}))\right) \,.
\end{align}

As a whole, there are 6 learnable linear projections and 2 LayerNorm modules in each ViT block. It is worth noting that ViT only extracts the classification token in the last block as the final representation, which is a different design from the practice in CNNs (i.e., global average pooling of all patch embeddings).

\subsection{Calculating the importance scores}  \label{sec:Calculate}

Since our goal is to compress ViT and also to preserve the consistency of token representations, structured channel pruning stands out as a reasonable choice. Firstly, we focus on calculating the importance scores of all channels (i.e., the $D$ dimensions). Our goal is to minimize the information loss of the last layer after pruning channels. Generally speaking, we divide ViTs into several uncorrelated components and evaluate the performance change after removing each channel in every component.

\begin{figure}[t]
  \centering  
  \includegraphics[width=0.45\textwidth]{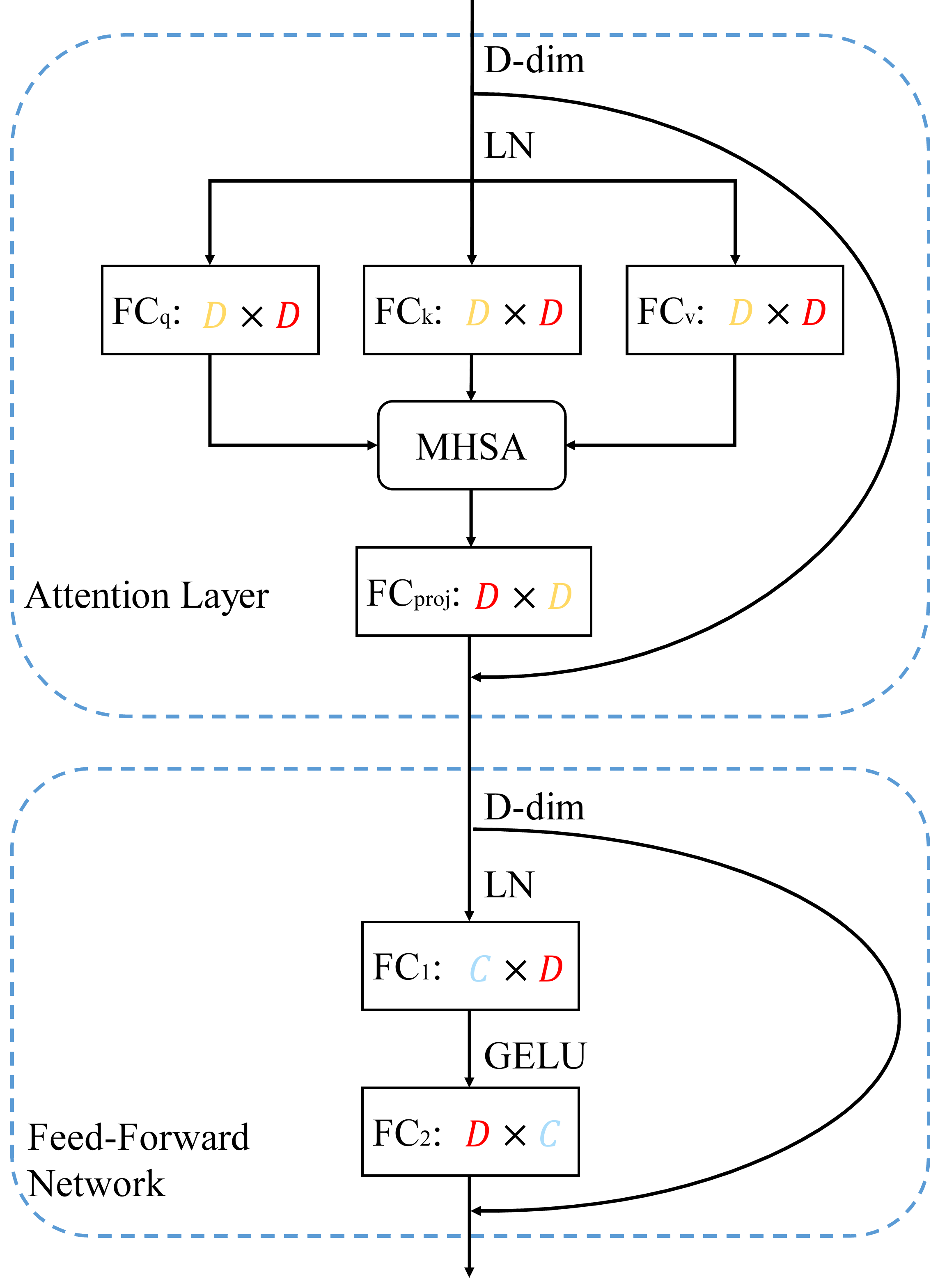}
  \caption{Illustration of three uncorrelated components in a ViT block. Our method prunes not only the channels inside the attention layer (the numbers shown in gold color) and the FFN (the numbers shown in blue), but also the channels across shortcut connections (the numbers shown in red). The two numbers in each rectangle represent the values of output and input channels of the linear projections, respectively.}
  \label{pipleine_vit}
\end{figure}

Let us take one ViT block as an example. As shown in Figure~\ref{pipleine_vit}, we divide the block into three irrelevant structural components, namely:
\squishlist
    \item Component 1: The shortcut connections that chain representations across \emph{all} blocks, i.e., the input channels of $\mathrm{FC}_q$, $\mathrm{FC}_k$, $\mathrm{FC}_v$ and $\mathrm{FC}_1$, the output channels of $\mathrm{FC}_{proj}$ and $\mathrm{FC}_2$, and the two $\mathrm{LN}$ layers;
    \item Component 2: The attention embedding filters inside the attention layer in \emph{every} block, i.e., the input channels of $\mathrm{FC}_{proj}$ and the output channels of $\mathrm{FC}_q$, $\mathrm{FC}_k$ and $\mathrm{FC}_v$;
    \item Component 3: The FFN inter-layer filters in \emph{every} block, i.e., the input channels of $\mathrm{FC}_2$ and the output channels of $\mathrm{FC}_1$.
\squishend

In particular, one crucial difficulty arises out of this structure. Because of the existence of shortcut connections, the following numbers \emph{must be the same in the entire network} after pruning: the number of channels in all components linked by the shortcut connections, the patch embedding dimensionality, and the final FC layer dimensionality. To make sure the validity of all summation and concatenation operations, the dimensionality of the classification token and the position embedding must be the same as this number, too.

To measure the channel importance, we randomly select 2000 images from the training dataset to establish a proxy dataset $\mathcal{D}$. We will extract the output logits on $\mathcal{D}$ and evaluate the performance change before/after removing a specific channel, e.g., when calculating the $k$th channel's importance score of the component 3, we will remove the $k$th input channel of $\mathrm{FC}_2$ and the $k$th output channel of $\mathrm{FC}_1$. Inspired by CURL \cite{luo2020neural}, the score is calculated by the KL-Divergence between two models with and without this particular channel, i.e.:
\begin{equation}
    s = \sum_{i \in \mathcal{D}} D_{\mathrm{KL}} ( q_i || p_i),
\end{equation}
where $i$ enumerates a sample from $\mathcal{D}$,  $p_i$ is the output of model without this particular channel, and $q_i$ is the output of the original model. The larger the score $s$, the more important this channel is. Note that because MHSA contains a reshaping operation, when calculating the importance scores of the component 2, we will mask the target channel as 0 instead of removing it, otherwise the reshaping operation will become invalid. 

Our method can evaluate the impact of all filters, and after generating all importance scores, we can produce more than one sub-network from a well-trained large model. The set of channels to be removed depend on these scores and different pre-set compression goals, which is flexible.

\subsection{Pruning and fine-tuning}  \label{sec:Fine-tuning} 

Given the original model and every channel's importance score, we first generate the sub-model candidate. In \emph{each} component and \emph{each} block, based on the pre-set compression ratio, we \emph{independently} rank the importance scores and delete those less important ones. Note that for simplicity, we use the same compression ratio in all components and all blocks.

In particular, the attention layer in ViT is benefited from the multi-head attention mechanism, which captures richer information by using multiple different heads. However, it also brings difficulties to compress the component 2. Therefore, we specifically design a simple but effective method to prune the multi-heads. In detail, given a $D_b$ dimensional attention layer with $h_b$ heads, we need to prune it into $D_t$ dimensions with $h_t$ heads. Note that every head contains the same number of dimensions across all attention layers and $h_b$ is divisible by $h_t$. The two settings are common in the transformer-based models. When pruning multi-head attention, we first merge every $h_b/h_t$ heads into 1 head, hence the attention layers will each have $h_t$ heads with $D_b$ dimensions. Then, we remove $\frac{D_b-D_t}{h_t}$ dimensions from every head, and the remaining module is the attention layer we want.

Lastly, we fine-tune the sub-model with all training samples, and use the original model as the teacher to distill the sub-model. Different from the training strategy of DeiT, we use the classic soft distillation to measure the training loss:
\begin{equation}
    \mathcal{L} = \mathcal{L}_{\mathrm{CE}}(y,p) + \alpha \mathcal{L}_{\mathrm{KL}}(q,p) \,, \label{eq:KD}
\end{equation}
where $p$ is the output probability (after softmax) of the sub-model, $q$ is the output probability of the teacher (i.e., the original model), and $y$ is the true label. $\mathcal{L}_{\mathrm{CE}}$ and  $\mathcal{L}_{\mathrm{KL}}$ denote the cross-entropy loss and the KL-divergence loss, respectively. Later, we will show that it is a better loss criterion than hard distillation when compressing ViTs, which is the opposite of the observation from DeiT.

\subsection{Pruning blocks} \label{sec:Pruning_Blocks}

Finally, we propose a new way to prune blocks in transformer-based models. More precisely, when pruning width, we first calculate all channels' important scores and then prune all redundant channels at once. Different from this pipeline, we construct a progressive method to compress blocks.
\squishlist
    \item First, for each block, calculate the KL-Divergence with \& without it;
    \item Second, remove the least important block and reserve the pruned model;
    \item Third, if the pruned model has not reached the compression target, then use the first two steps to prune one more block; otherwise we obtain a sub-model with fewer blocks and fine-tune it. 
\squishend

Note that the output of one block is its next block's input, so once we remove a block, the importance of every remaining block will change, and blocks adjacent to the deleted block will become more important. Therefore, we prune one block at a time. Later we will show this progressive block pruning method is better than pruning many blocks at once.
 
Unlike CNNs, given a ViT model with $L$ blocks, if we remove the FFN of the $k$th ($k\in \{1,2,\dots,L-1\}$) block and the attention layer of the next block, the remaining model is still guaranteed to be a legal ViT with $L-1$ blocks. Therefore, when compressing blocks, we actually compute $2L-1$ instead of $L$ importance scores. The $L-1$ new \emph{hybrid} block pruning candidates we propose here are \emph{novel and useful}.

\section{Experiments}

We now evaluate the performance of our UP-ViTs. We first prune DeiT-B and PVTv2-B2, and test the effectiveness on ImageNet-1k. In addition, more results on downstream small-scale classification and object detection datasets will be presented. We will show that UP-ViTs achieve better or comparable accuracies on these tasks. Finally, we will end this section with several analyses. All the experiments are conducted with PyTorch.

\subsection{Datasets and metrics}

\textbf{Classification.} The ImageNet-1k~\cite{deng2009imagenet} dataset consists of 1.28 million training and 50000 validation images. Those images have various spatial resolutions and come from 1000 different categories. ImageNet-1k is usually used as the benchmark for model pruning.

Besides ImageNet-1k, we also evaluate UP-ViTs on downstream small-scale datasets. Table~\ref{smalldatasets} summarizes the information of these datasets, including training and validation sizes, and the number of categories. 

\begin{table}
\centering
  \caption{Summary of small-scale datasets.}
  \small
\begin{tabular}{lrrr}
\bottomrule[1pt]
  Dataset    & Train        & Validation  & Categories \\ \hline
  CIFAR-100~\cite{krizhevsky2009learning} & 50000 & 10000         & 100\\ 
  CUB-200~\cite{welinder2010caltech}    & 5994 &	5794 &	200\\ 
  Cars~\cite{Krause_3DRR} & 8144 &	8041 &	196          \\ 
  Aircraft~\cite{maji13fine-grained}& 6667 &	3333 &	100           \\ 
  Indoor67~\cite{quattoni2009recognizing} & 5360&		1340 &	67 \\ 
  Pets~\cite{parkhi12a} & 3680&		3669&	37 \\ 
  DTD~\cite{cimpoi14describing}  & 3760&	1880&	47 \\ 
  iNaturalist-2019~\cite{van2018inaturalist} & 265213& 3030 & 1010 \\ \toprule[1pt]
\end{tabular}
\label{smalldatasets}
\end{table}

\textbf{Objection Detection.} We evaluate object detection performance on the MS COCO2017~\cite{lin2014microsoft} and the Pascal VOC07+12 \cite{everingham2010pascal} datasets. MS COCO 2017 contains 80 categories with 118K training and 5K validation images, respectively. Pascal VOC07+12 has 20 classes. Specifically, VOC07 contains a train-val set of 5011 images and a test set of 4952 images, and VOC12 contains a train-val set of 11540 images. We use mean Average Precision (mAP) to measure the detection accuracy. 

\subsection{Pruning DeiT-B on ImageNet-1k}\label{Pruning_DeiT}

We first prune DeiT-B on ImageNet-1k to DeiT-S \& T, and to further show the validity of our new block pruning strategy, we also prune blocks on the basis of the pruned DeiT-T.
 
\textbf{Implementation details.} First we prune DeiT-B into UP-DeiT-S~\&~T. Note that our compressed UP-DeiT-S~\&~T share the same structure as the original DeiT-S (for small) \& DeiT-T (for tiny). After all channel scores were calculated, we removed all redundant channels at once, then fine-tuned it with knowledge distillation in 200 epochs. We initialized the learning rate as 3e-4 and set $\alpha$ in Equation~\ref{eq:KD} as 0.5 when fine-tuning UP-DeiT-S. During fine-tuning UP-DeiT-T, the learning rate and $\alpha$ were 1e-3 and 0.2, respectively. After obtaining the compressed UP-DeiT-T model, we continue to delete 2 more blocks and fine-tune the sub-model. During fine-tuning, we set the learning rate as 3e-4 and trained 50 epochs with knowledge distillation with $\alpha$ being 0.4.

In the above experiments, we used the AdamW~\cite{loshchilov2018decoupled} optimizer and the cosine decay schedule. The weight decay was 1e-3 and the mini-batch size was 256. Random cropping, random horizontal flipping, color jittering and CutMix~\cite{yun2019cutmix} were applied as data augmentations. The mixing ratio of CutMix was 0.5. In particular, unlike DeiT, we did \emph{not} use mixup~\cite{zhang2017mixup}, random erasing~\cite{zhong2020random} or Rand-Augment~\cite{cubuk2020randaugment} in any of our experiments.

\textbf{Results.} Table~\ref{pruningdeit} shows the pruning results of DeiT-B. We tested model accuracy on the ImageNet-1k validation dataset. During testing, the shorter side was resized as 256 by bilinear interpolation and then we cropped the 224$\times$224 image patch in the center. The accuracy of the last epoch was reported. We also list the throughput in a TITAN Xp GPU with a fixed 32 mini-batch size.

\begin{table}[tbp]
  \centering
    \caption{Results of pruning DeiT-B in ImageNet-1k. UP-DeiT-S \& T and UP-DeiT-T-10 are our pruned model with 12 and 10 blocks respectively. Note that we denote the method proposed by~\cite{jia2021efficient} as MD-DeiT (manifold distillation DeiT). $^\star$  methods denote the results copied from the original paper.}
    \small
  \begin{tabular}{lrrr}
    \bottomrule[1pt]
  Model    & Throughput  & \#Param. &  Top-1 Acc. \\ \hline
  DeiT-B~\cite{Touvron2021deit}   &   199.2  & 86.6M  & 81.84\%   \\ \hline
  DeiT-S~\cite{Touvron2021deit}    &   603.1  & 22.1M  & 79.85\%   \\ 
  T2T-ViT-14~\cite{Yuan2021Tokens} &   456.6   & 21.5M  & 81.38\%   \\
  Swin-T~\cite{liu2021Swin}   &   384.5  & 28.5M  & 81.17\%   \\
  PVT-S~\cite{wang2021pyramid}  &  389.9 & 24.5M & 79.79\%   \\
  VTP$^\star$~\cite{zhu2021visual}       &   -   & 48.0M  & 80.70\%   \\
  NViT-S$^\star$~\cite{yang2021nvit}  & -    & 23.0M  & 81.22\%  \\
  Evo-ViT~\cite{xu2021evo}  & 305.7    &   87.3M   & 81.11\%   \\
  MD-DeiT-S$^\star$~\cite{jia2021efficient} &  603.1  & 22.1M  & 81.48\%  \\
  UP-DeiT-S    &   603.1  & 22.1M  & \textbf{81.56\%}  \\ \hline
  DeiT-T~\cite{Touvron2021deit}   &     1408.5   & 5.7M  & 72.20\%          \\
  T2T-ViT-10~\cite{Yuan2021Tokens} &     857.9  & 5.9M  & 75.00\%    \\
  PVT-T~\cite{wang2021pyramid}  &  691.7 & 13.2M & 75.00\%  \\
  NViT-T$^\star$~\cite{yang2021nvit}  & -    & 6.4M  & 73.91\%  \\
  MD-DeiT-T$^\star$~\cite{jia2021efficient} &  1408.5   & 5.7M  & 75.06\%  \\
  UP-DeiT-T    &   1408.5    & 5.7M   &  \textbf{75.79\%} \\ 
  UP-DeiT-T-10 &   1674.2    & 4.8M   &  73.97\%  \\
  \toprule[1pt]
  \end{tabular}
  \label{pruningdeit}
  \end{table}

Performance comparison between DeiT and UP-DeiT proves the effectiveness of our framework. Besides this, UP-DeiT also \emph{consistently} outperforms previous state-of-the-art ViT variants, such as T2T-ViT, Swin Transformer and PVT. And compared with DeiT-T, the UP-DeiT-T-10 achieves 1.58\% higher accuracy with only 10 blocks. Furthermore, UP-DeiT-S achieves 81.53\% accuracy with 3.03x (603.1 / 199.2) acceleration. This is better than Evo-ViT (state-of-the-art token-slimming method), which achieves 81.11\% accuracy with 1.53x acceleration.

\textbf{Transferring to small datasets.} We also tested the compressed models' transferring ability. We adopted mini-batch size 256 and learning rate 1e-3 when fine-tuning the DeiT-T model, and for the compressed models the learning rate was 3e-3. During fine-tuning, we applied the AdamW optimizer and the CutMix augmentation strategy.

\begin{table}
\centering
  \caption{Accuracy (\%) on different small-scale classification datasets. We train models with 50 epochs. }
  \small
\begin{tabular}{lccc}
  \bottomrule[1pt]
  Dataset    & DeiT-T        & UP-DeiT-T  & UP-DeiT-T-10 \\ \hline
  CIFAR-100  & 85.01 & \textbf{86.52}         & 85.98\\ 
  CUB-200    & 76.75 &\textbf{81.34}  &  80.93 \\ 
  Cars       & 85.24 & \textbf{87.34} & 86.27          \\ 
  Aircraft   & 72.58 &75.08 & \textbf{75.81}          \\ 
  Indoor67   & 74.70& \textbf{78.21} & 77.99 \\ 
  Pets       & 88.93 &\textbf{92.20} & 91.69 \\ 
  DTD        & 71.28 &\textbf{73.83} & 72.50 \\ 
  iNaturalist-2019 & 69.54&\textbf{70.86} & 69.17 \\ 
  \toprule[1pt]
  \end{tabular}
  \label{transfersmalldataset}
\end{table}

Table~\ref{transfersmalldataset} shows the results. The pruned model UP-DeiT-T \emph{always} outperforms DeiT-T on all 8 datasets, which indicates that our UP-ViTs method boosts the generalization of DeiT significantly. Also, except on the iNaturalist-2019 dataset, the 10-block UP-DeiT-T-10 outperforms the 12-block DeiT-T on other datasets.

\subsection{Pruning PVTv2 on ImageNet-1k} \label{Pruning_PVTv2}

In this section, we prune PVTv2 on ImageNet-1k. Our goal is to demonstrate that our method applies not only to pure transformer structures, but also to those variants that combine MHSA and convolution layers. We adopted two tasks to show our method's effectiveness, i.e., only prune depth, and prune both depth and width.
 
\textbf{Implementation details.} We conducted two experiments for comparison. In the first one, we deleted blocks and pruned PVTv2-B2 into UP-PVTv2-B1; in the second, we directly pruned PVTv2-B2 into UP-PVTv2-B0. Note that the structures of UP-PVTv2-B1 \& B0 are exactly the same as those of PVTv2-B1 \& B0, respectively.  It is worth noting that PVTv2 introduces a convolutional layer into the attention module to reduce the computational cost, so we added one extra component when calculating the importance scores.
 
During training, we followed the data augmentation and learning rate setting of pruning DeiT-B. In particular, when pruning PVTv2-B2 to UP-PVTv2-B0, we used a learning rate of 1e-3 and an $\alpha$ of 0.2. The sub-model was fine-tuned for 100 epochs. When we pruned PVTv2-B2 to UP-PVTv2-B1, the learning rate were initialized as 3e-4 and we trained 50 epochs. Especially, we found that directly distilling the features in the penultimate layer (after the global average pooling and before the final classifier) is advantageous, so we applied the MSE loss instead of the KL-divergence loss as the distillation loss. The $\alpha$ was set to 20.

\begin{table}
  \centering
    \caption{Results of pruning PVTv2 on ImageNet-1k. We report the accuracy of the last epoch on ImageNet-1k validation. UP-PVTv2-B1 \& B0 are our compressed models. }
	\small
 \begin{tabular}{c|ccc}
  \bottomrule[1pt]
  Method    & Throughput  & \#Param.  & Top-1 Acc. \\ \hline
  PVTv2-B2     &  238.0 & 25.36M  & 82.08\%  \\ \hline
  PVTv2-B1  &  \multicolumn{1}{c}{\multirow{2}{*}{ 407.2 }}   & \multicolumn{1}{c}{\multirow{2}{*}{ 14.01M }}  & 78.62\%         \\
  UP-PVTv2-B1 &      &   & \textbf{79.48\%} \\ \hline
  PVTv2-B0 &   \multicolumn{1}{c}{\multirow{2}{*}{ 739.1 }}  &  \multicolumn{1}{c}{\multirow{2}{*}{ 3.67M  }}  &  70.47\%    \\ 
  UP-PVTv2-B0  &     & &  \textbf{75.30\%}   \\
    \toprule[1pt]
\end{tabular}
\label{pruningPVTv2}
\end{table}

\textbf{Results.} Table~\ref{pruningPVTv2} shows the results on PVTv2, and similar to the previous experiments, our UP-PVTv2 models obtain significantly better results when compared with the original models. 

\textbf{Transferring to object detection.} To further validate the effectiveness of our method on a larger object detection dataset, we investigate UP-PVTv2's performances on object detection with Mask R-CNN~\cite{he2017mask} and RetinaNet~\cite{lin2017focal}. First, we adopted both PVTv2-B0/B1 and our  UP-PVTv2-B0/B1 as backbones of RetinaNet on Pascal VOC07+12.

\begin{table}
  \centering
    \caption{mAP of  UP-PVTv2 on Pascal VOC07 test. We trained RetinaNet on the Pascal VOC07+12 train-val dataset. }
	\small
 \begin{tabular}{c|ccc}
  \bottomrule[1pt]
  
Backbone   & \#Param. & FLOPs & mAP   \\ \hline
PVTv2-B1   & \multicolumn{1}{c}{\multirow{2}{*}{22.5M}}  & \multicolumn{1}{c}{\multirow{2}{*}{108.8G}}  &  81.9 \\
UP-PVTv2-B1 &    & &  \textbf{83.0}\\     \hline
PVTv2-B0   & \multicolumn{1}{c}{\multirow{2}{*}{11.7M}}  & \multicolumn{1}{c}{\multirow{2}{*}{85.0G}} & 80.5 \\
UP-PVTv2-B0 &   & & \textbf{82.0} \\
  \toprule[1pt]
\end{tabular}
\label{RetinaNetVoC}
\end{table}

As reported in Table~\ref{RetinaNetVoC}, the compressed UP-PVTv2 models significantly outperform original models on Pascal VOC detection. For example, UP-PVTv2-B0 achieves 82.0 mAP, which surpasses the original PVTv2-B0 model by 1.5 AP points, and is on par with the performance of the original PVTv2-B1.

\begin{table*}
  \centering
    \caption{mAP of (UP-)PVTv2-B1 on the MS COCO2017 validation dataset. }
    \small
\begin{tabular}{c|ccc|ccc|ccc}
  \bottomrule[1pt]
Backbone   & Methods &\#Param.& FLOPs& AP &AP$_{50}$ &AP$_{75}$ & AP$_{S}$  &  AP$_{M}$ &AP$_{L}$    \\ \hline
PVTv2-B1   &   \multicolumn{1}{c}{\multirow{2}{*}{RetinaNet 1x}} &  \multicolumn{1}{c}{\multirow{2}{*}{23.75M}}  &  \multicolumn{1}{c|}{\multirow{2}{*}{235.9G}}& 40.5 & 61.0 & 43.3 & 24.0 & 43.9 & 53.1  \\
UP-PVTv2-B1 &  & &   & \textbf{41.1}& \textbf{62.0} & \textbf{43.5}  & \textbf{25.1}& \textbf{44.2} &\textbf{54.6} \\  \hline
PVTv2-B1  &  \multicolumn{1}{c}{\multirow{2}{*}{Mask R-CNN 1x}}  &  \multicolumn{1}{c}{\multirow{2}{*}{33.66M}}  &  \multicolumn{1}{c|}{\multirow{2}{*}{252.2G}} & 40.8 & 63.1 & 44.5 & 25.3 & 44.5 & 53.0  \\
UP-PVTv2-B1 &  & &  & \textbf{41.6} &  \textbf{64.1} &  \textbf{45.5} &  \textbf{25.9} &  \textbf{45.0}  & \textbf{54.0}  \\
  \toprule[1pt]
\end{tabular}
\label{coco}
\end{table*}

We also investigated the performance of (UP-)PVTv2-B1 backbone on both one-stage and two-stage object detectors, Mask R-CNN and RetinaNet. From the results shown in Table~\ref{coco} on the larger MS-COCO dataset, our method significantly outperforms original PVTv2-B1.

\subsection{Analyses}

To explore the impact of different modules of our method, we performed three analyses in this section.

\subsubsection{Pruning multi-head self-attention}

We first evaluate the influence of our attention pruning strategy when pruning component 2 by conducting an experiment of pruning DeiT-B into UP-DeiT-T. In particular, DeiT-B and UP-DeiT-T are similar in structure, but DeiT-B contains 768 dimensions with 12 heads in its attention layer, and UP-DeiT-T has 192 dimensions with only 3 heads. After calculating the attention importance score of DeiT-B, we compare three strategies of pruning the component 2:
\squishlist
    \item Strategy 1: Our strategy of pruning the attention layer in Section~\ref{sec:Fine-tuning}. For the 12 heads, we remove 192 dimensions from every four heads.
    \item Strategy 2: Do not consider the structure difference and delete 576 dimensions across all heads;
    \item Strategy 3: For the 12 heads, we remove 48 dimensions for each attention head;
\squishend

After deleting those redundant dimensions, we merged the remaining filters into 3 heads and extracted three different UP-DeiT-T sub-models. We sampled the original ImageNet-1k to a smaller subset with one-tenth of the total images, which is named SImageNet-1k. Then we fine-tuned the three sub-models 50 epochs on SImageNet-1k and tested on the original ImageNet-1k validation set. The other training settings were the same as those Section~\ref{Pruning_DeiT}.  We show the results in Table~\ref{samplingattention}, and strategy 1 achieves the highest accuracy.


\begin{table}
  \centering
    \caption{Results on the ImageNet-1k validation set with different attention pruning strategies.}
    \small
 \begin{tabular}{cccc}
  \bottomrule[1pt]
   & Strategy 1 & Strategy 2 & Strategy 3 \\  \hline
Accuracy     & \textbf{48.51\%}& 48.45\% & 48.05\% \\
\toprule[1pt]
\end{tabular}
\label{samplingattention}
\end{table}

\subsubsection{Knowledge distillation}

We then evaluate the influence of knowledge distillation when fine-tuning UP-ViTs. For a fair comparison, we adopt the same training strategies as Section~\ref{Pruning_DeiT} and Section~\ref{Pruning_PVTv2}. We illustrate the results in Table~\ref{distillationstrategy}, which shows our method still obtains competitive accuracy when training without knowledge distillation, and UP-DeiT-T even has better performance without distillation. 
We further elongate the training process for random initialization, without distillation and with DeiT-B distillation for DeiT-T. Our pruned model are more accurate than model training from scratch with more epochs, which shows that our pruning strategy is better than directly training a light model from scratch. 
\begin{table}
  \centering
    \caption{Results on the ImageNet-1k validation set with \& without knowledge distillation.  }
    \small
 \begin{tabular}{cccc}
  \bottomrule[1pt]
Model                   & Distillation? &  Epochs & Top 1 Acc.  \\ \hline
\multicolumn{1}{c}{\multirow{2}{*}{DeiT-T }} & \CheckmarkBold & 500 & 74.54\%     \\
\multicolumn{1}{c}{}                     & \XSolidBrush & 500 & 72.65\%   \\ \hline
\multicolumn{1}{c}{\multirow{2}{*}{UP-DeiT-T}} & \CheckmarkBold & 200 &   75.79\%  \\
 \multicolumn{1}{c}{}                     & \XSolidBrush & 200 &   76.11\%\\\hline
 \multicolumn{1}{c}{\multirow{2}{*}{UP-PVTv2-B0}} & \CheckmarkBold & 100 & 75.30\%     \\
 \multicolumn{1}{c}{}                     & \XSolidBrush & 100 & 74.33\%   \\
 \toprule[1pt]
\end{tabular}
\label{distillationstrategy}
\end{table}

We also compare several different distillation strategies. Besides soft distillation, we consider the following two methods.

\textbf{Hard distillation.} Let $y_t = \arg\max(q)$ be the hard decision of the teacher, where $q$ is the teacher's output probabilities. The loss function of hard-label distillation is
\begin{equation}
  \mathcal{L}_{\mathrm{CE}}\left(p, y\right)+\alpha \mathcal{L}_{\mathrm{CE}}\left(p, y_t\right) \,.
 \end{equation}
 
\textbf{Soft + Patch distillation.} Inspired by LV-ViT \cite{jiang2021all}, we distill both the classification token and the patch tokens.  Assume the student's output patch tokens are $t_s^{1},\dots,t_s^{N}$ and the corresponding outputs of the teacher are $t_t^{1},\dots,t_t^{N}$. Note that the number of embedding dimensions in $t_s^{i}$ and $t_t^{i}$ are different, so we introduce a new linear projection $\mathrm{FC}_{token}$ into the loss function. The soft + patch distillation objective is: 
\begin{equation}
  \mathcal{L}_{\mathrm{CE}}\left(p, y\right) +   \alpha \mathcal{L}_{KL}(p,q) 
     + \beta \sum_{i=1}^{N}  \mathcal{L}_{\mathrm{MSE}}\left(\mathrm{FC}_{token}(t_s^{i}),t_t^{i}\right) \,.
  \end{equation}
  
Based on the three strategies, we investigate the performance of distilling UP-DeiT-T with DeiT-B. During fine-tuning, we set $\alpha$ of hard distillation as 2.0. When applying soft + patch distillation, we set $\alpha$ and $\beta$ as 0.2 and 1e-3, respectively. As the distillation results in Table~\ref{distillation} clearly show, soft distillation achieves the highest accuracy.

\begin{table}
  \centering
    \caption{Results on the ImageNet-1k validation set with 3 distillation strategies. We trained the models in SImageNet-1k with 50 epochs.}
	\small
 \begin{tabular}{c|ccc}
  \bottomrule[1pt]
Distillation   & Soft & Hard & Soft + Patch \\  \hline
Accuracy     & \textbf{51.51\%} & 48.95\% &  51.49\%\\
\toprule[1pt]
\end{tabular}
\label{distillation}
\end{table}

\subsubsection{Progressive vs. One-time pruning}

The last sets of experiments are about progressive vs. one-time pruning. We compare two pipelines during compressing DeiT-B. In the first pipeline, after calculating all importance scores, we remove all redundant channels at once and fine-tune the model. In the second pipeline, we divide the whole pruning process into 4 steps:
\squishlist
    \item Step 1: Calculate the importance scores of component 1 in DeiT-B. There are 768 channels and we throw away 576 of them and fine-tune the sub-model.
    \item Step 2: Continue to calculate the importance scores of component 2 per block and prune it. In particular, we throw away 48 dimensions for every attention head and keep 12 attention heads, then fine-tune the sub-model.
    \item Step 3: Similar to the second step, we prune component 3 and fine-tune the model. Especially, the dimensionalities of the FFN hidden layer are reduced from 3072 to 768.
    \item Step 4: Adjust the attention heads from 12 to 3 and fine-tune the model.
\squishend

Note that during fine-tuning, we apply knowledge distillation and the teacher model is always DeiT-B. The results are shown in Table~\ref{table9}. We can observe that the performance of progressive pruning (74.51\%) is not as good as pruning at once (75.79\%).

\begin{table}
  \centering
    \caption{Results of the progressive pipeline when pruning to DeiT-T on ImageNet-1k. In every step we fine-tuned model with 50 epochs. }
    \small
    \begin{tabular}{cccc}
      \bottomrule[1pt]
    Model   & Top-1 Acc.  & Throughput  & \#Param. \\ \hline
    DeiT-B & 81.84\%     &  58.1 & 86.57M   \\ \hline
    \phantom{+ }Step 1 & 79.90\%     &  139.4     &   21.69M       \\
    + Step 2 & 79.60\%     &   188.2    &  16.36M         \\
    + Step 3  &  75.05\%   &  296.7     &   5.72M   \\
    + Step 4  &  74.51\%  &    404.5   &   5.72M  \\
    \toprule[1pt]
    \end{tabular}
\label{table9}
\end{table}

We then investigate the performance of progressively pruning the component 1 of DeiT-B. We compare two pipelines. In the first, we remove 576 redundant filters at once after calculating the importance of all filters in component 1. In the second pipeline, we imitate the process of pruning blocks, and construct a loop that only throws away 192 filters in component 1 every time, and repeat this loop three times. The results are shown in Table~\ref{Components1}. The first pipeline clearly works better, which indicates pruning all redundant filters at once is most effective when compressing the width of ViTs.

\begin{table}
  \centering
    \caption{Results of two pipelines when pruning component 1 on SImageNet-1k. Pruning at once represents the first pipeline and pruning progressively is the second one. We fine-tuned models with 50 epochs. }
    \small
    \begin{tabular}{c|cc}
      \bottomrule[1pt]
    Component 1    & Pruning at once & Pruning progressively  \\ \hline
    Accuracy  &  \textbf{66.95\%} & 65.05\% \\ \hline

    \toprule[1pt]
    \end{tabular}
\label{Components1}
\end{table}

We also demonstrate the effectiveness of progressively pruning blocks. Based on the compressed UP-DeiT-T, we compare two strategies for generating UP-DeiT-T-10. In the first we directly remove the two least unimportant blocks at one time, and in the second we use our progressive block pruning strategy. Once we get the two sub-model candidates, we apply the same training policy to fine-tune model on ImageNet-1k and show the results in Table~\ref{pruningblock}. The progressive block pruning strategy obtains higher accuracy.

\begin{table}
  \centering
    \caption{Results of two pipelines when pruning UP-DeiT-T into UP-DeiT-T-10 on ImageNet-1k. We fine-tune models with 30 epochs and with knowledge distillation.}
    \small
    \begin{tabular}{c|cc}
      \bottomrule[1pt]
    Blocks    & Pruning at once  & Pruning progressively \\ \hline
    Accuracy   &  73.20\% & \textbf{73.78\%}\\ \hline
    \toprule[1pt]
    \end{tabular}
\label{pruningblock}
\end{table}

\section{Conclusion, Limitations, and Future Work}

In this paper, we proposed a novel method, UP-ViTs, for pruning ViTs in a unified manner. Our framework can prune all components in ViT and its variants, maintain the models' structure and generalize well into downstream tasks. Especially, UP-ViTs achieved state-of-the-art results when pruning various ViT backbones. Moreover, we studied the transferring ability of the compressed model and found that our UP-ViTs also had better performances than original ViTs.  

Although our framework can be applied to various ViT architectures, we have to manually identify the uncorrelated components existing in ViTs (cf. the appendix), hence how to automatically divide and prune the irrelevant components is an interesting future direction. Besides this, we focus on pruning heavy models into existing lightweight model architectures, such as compressing DeiT-B into DeiT-T. Therefore, continuing to prune ViTs to make them more attractive for embedded systems is another interesting direction. Transformer was originally designed to solve NLP problems and many previous works proposed various MHSA mechanisms, so in the future, we will further generalize our pruning framework into NLP tasks. 

{\small
\bibliographystyle{ieee_fullname}
\bibliography{egbib}

\begin{thebibliography}{10}\itemsep=-1pt

\bibitem{Behnke2020losing}
Maximiliana Behnke and Kenneth Heafield.
\newblock Losing heads in the lottery: Pruning transformer attention in neural
  machine translation.
\newblock In {\em Proceedings of the 2020 Conference on Empirical Methods in
  Natural Language Processing (EMNLP)}, pages 2664--2674, 2020.

\bibitem{carion2020end}
Nicolas Carion, Francisco Massa, Gabriel Synnaeve, Nicolas Usunier, Alexander
  Kirillov, and Sergey Zagoruyko.
\newblock End-to-end object detection with transformers.
\newblock In {\em The European Conference on Computer Vision (ECCV)}, volume
  12346 of {\em LNCS}, pages 213--229. Springer, 2020.

\bibitem{choromanski2021rethinking}
Krzysztof~Marcin Choromanski, Valerii Likhosherstov, David Dohan, Xingyou Song,
  Andreea Gane, Tamas Sarlos, Peter Hawkins, Jared~Quincy Davis, Afroz
  Mohiuddin, Lukasz Kaiser, David~Benjamin Belanger, Lucy~J Colwell, and Adrian
  Weller.
\newblock Rethinking attention with {Performers}.
\newblock In {\em International Conference on Learning Representations (ICLR)},
  2021.

\bibitem{cimpoi14describing}
M. Cimpoi, S. Maji, I. Kokkinos, S. Mohamed, , and A. Vedaldi.
\newblock Describing textures in the wild.
\newblock In {\em {The IEEE Conference on Computer Vision and Pattern
  Recognition (CVPR)}}, pages 3606--3613, 2014.

\bibitem{cubuk2020randaugment}
Ekin~D Cubuk, Barret Zoph, Jonathon Shlens, and Quoc~V Le.
\newblock {RandAugment: Practical automated data augmentation with a reduced
  search space}.
\newblock In {\em The IEEE/CVF Conference on Computer Vision and Pattern
  Recognition (CVPR)}, pages 702--703, 2020.

\bibitem{deng2009imagenet}
Jia Deng, Wei Dong, Richard Socher, Li-Jia Li, Kai Li, and Li Fei-Fei.
\newblock {ImageNet: A large-scale hierarchical image database}.
\newblock In {\em {The IEEE Conference on Computer Vision and Pattern
  Recognition (CVPR)}}, pages 248--255, 2009.

\bibitem{dosovitskiy2020vit}
Alexey Dosovitskiy, Lucas Beyer, Alexander Kolesnikov, Dirk Weissenborn,
  Xiaohua Zhai, Thomas Unterthiner, Mostafa Dehghani, Matthias Minderer, Georg
  Heigold, Sylvain Gelly, Jakob Uszkoreit, and Neil Houlsby.
\newblock An image is worth 16x16 words: Transformers for image recognition at
  scale.
\newblock In {\em International Conference on Learning Representations (ICLR)},
  2021.

\bibitem{everingham2010pascal}
Mark Everingham, Luc Van~Gool, Christopher~KI Williams, John Winn, and Andrew
  Zisserman.
\newblock {The pascal visual object classes (VOC) challenge}.
\newblock {\em International Journal of Computer Vision}, 88(2):303--338, 2010.

\bibitem{fan2019reducing}
Angela Fan, Edouard Grave, and Armand Joulin.
\newblock Reducing transformer depth on demand with structured dropout.
\newblock In {\em International Conference on Learning Representations (ICLR)},
  2019.

\bibitem{guo2019star}
Qipeng Guo, Xipeng Qiu, Pengfei Liu, Yunfan Shao, Xiangyang Xue, and Zheng
  Zhang.
\newblock {Star-Transformer}.
\newblock In {\em Proceedings of the 2019 Conference of the North American
  Chapter of the Association for Computational Linguistics: Human Language
  Technologies, Volume 1 (Long and Short Papers)}, pages 1315--1325, 2019.

\bibitem{he2017mask}
Kaiming He, Georgia Gkioxari, Piotr Doll{\'a}r, and Ross Girshick.
\newblock {Mask R-CNN}.
\newblock In {\em Proceedings of the IEEE International Conference on Computer
  Vision (ICCV)}, pages 2961--2969, 2017.

\bibitem{he2019filter}
Yang He, Ping Liu, Ziwei Wang, Zhilan Hu, and Yi Yang.
\newblock Filter pruning via geometric median for deep convolutional neural
  networks acceleration.
\newblock In {\em Proceedings of the IEEE/CVF Conference on Computer Vision and
  Pattern Recognition (CVPR)}, pages 4340--4349, 2019.

\bibitem{he2017channel}
Yihui He, Xiangyu Zhang, and Jian Sun.
\newblock Channel pruning for accelerating very deep neural networks.
\newblock In {\em Proceedings of the IEEE International Conference on Computer
  Vision (ICCV)}, pages 1389--1397, 2017.

\bibitem{Huang2016deep}
Gao Huang, Yu Sun, Zhuang Liu, Daniel Sedra, and Kilian~Q Weinberger.
\newblock Deep networks with stochastic depth.
\newblock In {\em The European Conference on Computer Vision (ECCV)}, volume
  9908 of {\em LNCS}, pages 646--661. Springer, 2016.

\bibitem{jia2021efficient}
Ding Jia, Kai Han, Yunhe Wang, Yehui Tang, Jianyuan Guo, Chao Zhang, and
  Dacheng Tao.
\newblock Efficient vision transformers via fine-grained manifold distillation.
\newblock {\em arXiv preprint arXiv:2107.01378}, 2021.

\bibitem{jiang2021all}
Zihang Jiang, Qibin Hou, Li Yuan, Daquan Zhou, Yujun Shi, Xiaojie Jin, Anran
  Wang, and Jiashi Feng.
\newblock All tokens matter: Token labeling for training better vision
  transformers.
\newblock {\em arXiv preprint arXiv:2104.10858}, 2021.

\bibitem{Krause_3DRR}
Krause Jonathan, Stark Michael, Deng Jia, and Li Fei-Fei.
\newblock 3d object representations for fine-grained categorization.
\newblock In {\em International IEEE Workshop on 3D Representation and
  Recognition}, pages 554--561, 2013.

\bibitem{katharopoulos2020transformers}
Angelos Katharopoulos, Apoorv Vyas, Nikolaos Pappas, and Fran{\c{c}}ois
  Fleuret.
\newblock Transformers are rnns: Fast autoregressive transformers with linear
  attention.
\newblock In {\em International Conference on Machine Learning}, pages
  5156--5165, 2020.

\bibitem{krizhevsky2009learning}
Alex Krizhevsky and Geoffrey Hinton.
\newblock Learning multiple layers of features from tiny images.
\newblock Technical report, University of Toronto, 2009.

\bibitem{li2016pruning}
Hao Li, Asim Kadav, Igor Durdanovic, Hanan Samet, and Hans~Peter Graf.
\newblock Pruning filters for efficient convnets.
\newblock In {\em International Conference on Learning Representations (ICLR)},
  2017.

\bibitem{li2020train}
Zhuohan Li, Eric Wallace, Sheng Shen, Kevin Lin, Kurt Keutzer, Dan Klein, and
  Joey Gonzalez.
\newblock Train big, then compress: Rethinking model size for efficient
  training and inference of transformers.
\newblock In {\em International Conference on Machine Learning}, pages
  5958--5968, 2020.

\bibitem{lin2017focal}
Tsung-Yi Lin, Priya Goyal, Ross Girshick, Kaiming He, and Piotr Doll{\'a}r.
\newblock Focal loss for dense object detection.
\newblock In {\em Proceedings of the IEEE International Conference on Computer
  Vision (ICCV)}, pages 2980--2988, 2017.

\bibitem{lin2014microsoft}
Tsung-Yi Lin, Michael Maire, Serge Belongie, James Hays, Pietro Perona, Deva
  Ramanan, Piotr Doll{\'a}r, and C~Lawrence Zitnick.
\newblock {Microsoft COCO}: Common objects in context.
\newblock In {\em The European Conference on Computer Vision (ECCV)}, volume
  8693 of {\em LNCS}, pages 740--755. Springer, 2014.

\bibitem{liu2021Swin}
Ze Liu, Yutong Lin, Yue Cao, Han Hu, Yixuan Wei, Zheng Zhang, Stephen Lin, and
  Baining Guo.
\newblock {Swin Transformer}: Hierarchical vision transformer using shifted
  windows.
\newblock In {\em Proceedings of the IEEE/CVF International Conference on
  Computer Vision (ICCV)}, pages 10012--10022, 2021.

\bibitem{loshchilov2018decoupled}
Ilya Loshchilov and Frank Hutter.
\newblock Decoupled weight decay regularization.
\newblock In {\em International Conference on Learning Representations (ICLR)},
  2018.

\bibitem{luo2020neural}
Jian-Hao Luo and Jianxin Wu.
\newblock Neural network pruning with residual-connections and limited-data.
\newblock In {\em Proceedings of the IEEE/CVF Conference on Computer Vision and
  Pattern Recognition (CVPR)}, pages 1458--1467, 2020.

\bibitem{luo2017thinet}
Jian-Hao Luo, Jianxin Wu, and Weiyao Lin.
\newblock Thinet: A filter level pruning method for deep neural network
  compression.
\newblock In {\em Proceedings of the IEEE International Conference on Computer
  Vision (ICCV)}, pages 5058--5066, 2017.

\bibitem{maji13fine-grained}
S. Maji, J. Kannala, E. Rahtu, M. Blaschko, and A. Vedaldi.
\newblock Fine-grained visual classification of aircraft.
\newblock {\em arXiv preprint arXiv:1306.5151}, 2013.

\bibitem{marin2021token}
Dmitrii Marin, Jen-Hao~Rick Chang, Anurag Ranjan, Anish Prabhu, Mohammad
  Rastegari, and Oncel Tuzel.
\newblock Token pooling in vision transformers.
\newblock {\em arXiv preprint arXiv:2110.03860}, 2021.

\bibitem{parkhi12a}
Omkar~M. Parkhi, Andrea Vedaldi, Andrew Zisserman, and C.~V. Jawahar.
\newblock Cats and dogs.
\newblock In {\em {The IEEE Conference on Computer Vision and Pattern
  Recognition (CVPR)}}, pages 3498--3505, 2012.

\bibitem{quattoni2009recognizing}
Ariadna Quattoni and Antonio Torralba.
\newblock Recognizing indoor scenes.
\newblock In {\em {The IEEE Conference on Computer Vision and Pattern
  Recognition (CVPR)}}, pages 413--420, 2009.

\bibitem{tang2021patch}
Yehui Tang, Kai Han, Yunhe Wang, Chang Xu, Jianyuan Guo, Chao Xu, and Dacheng
  Tao.
\newblock Patch slimming for efficient vision transformers.
\newblock {\em arXiv preprint arXiv:2106.02852}, 2021.

\bibitem{Touvron2021deit}
Hugo Touvron, Matthieu Cord, Matthijs Douze, Francisco Massa, Alexandre
  Sablayrolles, and Herve Jegou.
\newblock Training data-efficient image transformers \& distillation through
  attention.
\newblock In {\em International Conference on Machine Learning}, pages
  10347--10357, 2021.

\bibitem{van2018inaturalist}
Grant Van~Horn, Oisin Mac~Aodha, Yang Song, Yin Cui, Chen Sun, Alex Shepard,
  Hartwig Adam, Pietro Perona, and Serge Belongie.
\newblock The inaturalist species classification and detection dataset.
\newblock In {\em {The IEEE Conference on Computer Vision and Pattern
  Recognition (CVPR)}}, pages 8769--8778, 2018.

\bibitem{Vaswani2017Attention}
Ashish Vaswani, Noam Shazeer, Niki Parmar, Jakob Uszkoreit, Llion Jones,
  Aidan~N Gomez, \L~ukasz Kaiser, and Illia Polosukhin.
\newblock Attention is all you need.
\newblock In I. Guyon, U.~V. Luxburg, S. Bengio, H. Wallach, R. Fergus, S.
  Vishwanathan, and R. Garnett, editors, {\em Advances in Neural Information
  Processing Systems}, volume~30, pages 5998--6008, 2017.

\bibitem{welinder2010caltech}
Catherine Wah, Steve Branson, Peter Welinder, Pietro Perona, and Serge
  Belongie.
\newblock {The Caltech-UCSD Birds-200-2011 Dataset}.
\newblock Technical Report CNS-TR-2011-001, California Institute of Technology,
  2011.

\bibitem{pvtv2}
Wenhai Wang, Enze Xie, Xiang Li, Deng-Ping Fan, Kaitao Song, Ding Liang, Tong
  Lu, Ping Luo, and Ling Shao.
\newblock {PVTv2: Improved} baselines with pyramid vision transformer.
\newblock {\em arXiv preprint arXiv:2106.13797}, 2021.

\bibitem{wang2021pyramid}
Wenhai Wang, Enze Xie, Xiang Li, Deng-Ping Fan, Kaitao Song, Ding Liang, Tong
  Lu, Ping Luo, and Ling Shao.
\newblock {Pyramid vision transformer: A versatile backbone for dense
  prediction without convolutions}.
\newblock In {\em Proceedings of the IEEE/CVF International Conference on
  Computer Vision (ICCV)}, pages 568--578, 2021.

\bibitem{wang2021end}
Yuqing Wang, Zhaoliang Xu, Xinlong Wang, Chunhua Shen, Baoshan Cheng, Hao Shen,
  and Huaxia Xia.
\newblock End-to-end video instance segmentation with transformers.
\newblock In {\em {The IEEE/CVF Conference on Computer Vision and Pattern
  Recognition (CVPR)}}, pages 8741--8750, 2021.

\bibitem{xu2021evo}
Yifan Xu, Zhijie Zhang, Mengdan Zhang, Kekai Sheng, Ke Li, Weiming Dong, Liqing
  Zhang, Changsheng Xu, and Xing Sun.
\newblock {Evo-ViT: Slow-Fast} token evolution for dynamic vision transformer.
\newblock {\em arXiv preprint arXiv:2108.01390}, 2021.

\bibitem{yang2021nvit}
Huanrui Yang, Hongxu Yin, Pavlo Molchanov, Hai Li, and Jan Kautz.
\newblock {NViT}: Vision transformer compression and parameter redistribution.
\newblock {\em arXiv preprint arXiv:2110.04869}, 2021.

\bibitem{Yuan2021Tokens}
Li Yuan, Yunpeng Chen, Tao Wang, Weihao Yu, Yujun Shi, Zi-Hang Jiang,
  Francis~E.H. Tay, Jiashi Feng, and Shuicheng Yan.
\newblock {Tokens-to-Token ViT: Training} vision transformers from scratch on
  imagenet.
\newblock In {\em Proceedings of the IEEE/CVF International Conference on
  Computer Vision (ICCV)}, pages 558--567, 2021.

\bibitem{yun2019cutmix}
Sangdoo Yun, Dongyoon Han, Seong~Joon Oh, Sanghyuk Chun, Junsuk Choe, and
  Youngjoon Yoo.
\newblock Cutmix: Regularization strategy to train strong classifiers with
  localizable features.
\newblock In {\em Proceedings of the IEEE/CVF International Conference on
  Computer Vision (ICCV)}, pages 6023--6032, 2019.

\bibitem{zhang2017mixup}
Hongyi Zhang, Moustapha Cisse, Yann~N. Dauphin, and David Lopez-Paz.
\newblock {Mixup: Beyond Empirical Risk Minimization}.
\newblock In {\em International Conference on Learning Representations (ICLR)},
  2018.

\bibitem{zhong2020random}
Zhun Zhong, Liang Zheng, Guoliang Kang, Shaozi Li, and Yi Yang.
\newblock Random erasing data augmentation.
\newblock In {\em Proceedings of the AAAI Conference on Artificial
  Intelligence}, pages 13001--13008, 2020.

\bibitem{zhu2021visual}
Mingjian Zhu, Kai Han, Yehui Tang, and Yunhe Wang.
\newblock Visual transformer pruning.
\newblock In {\em KDD 2021 Workshop on Model Mining}, 2021.

\end{thebibliography}
}

\newpage
\begin{appendix}
\section{The components of PVTv2}

PVTv2 proposes the overlapping patch embedding module to divide a model into four stages. It also introduces a convolution layer and a $3 \times 3$ depth-wise convolution into the attention layer and FFN, respectively. As shown in Figure~\ref{pipleine_pvtv2}, we demonstrate four uncorrelated components in a PVTv2 block:

\begin{figure}[htbp]
  \centering  
  \includegraphics[width=0.45\textwidth]{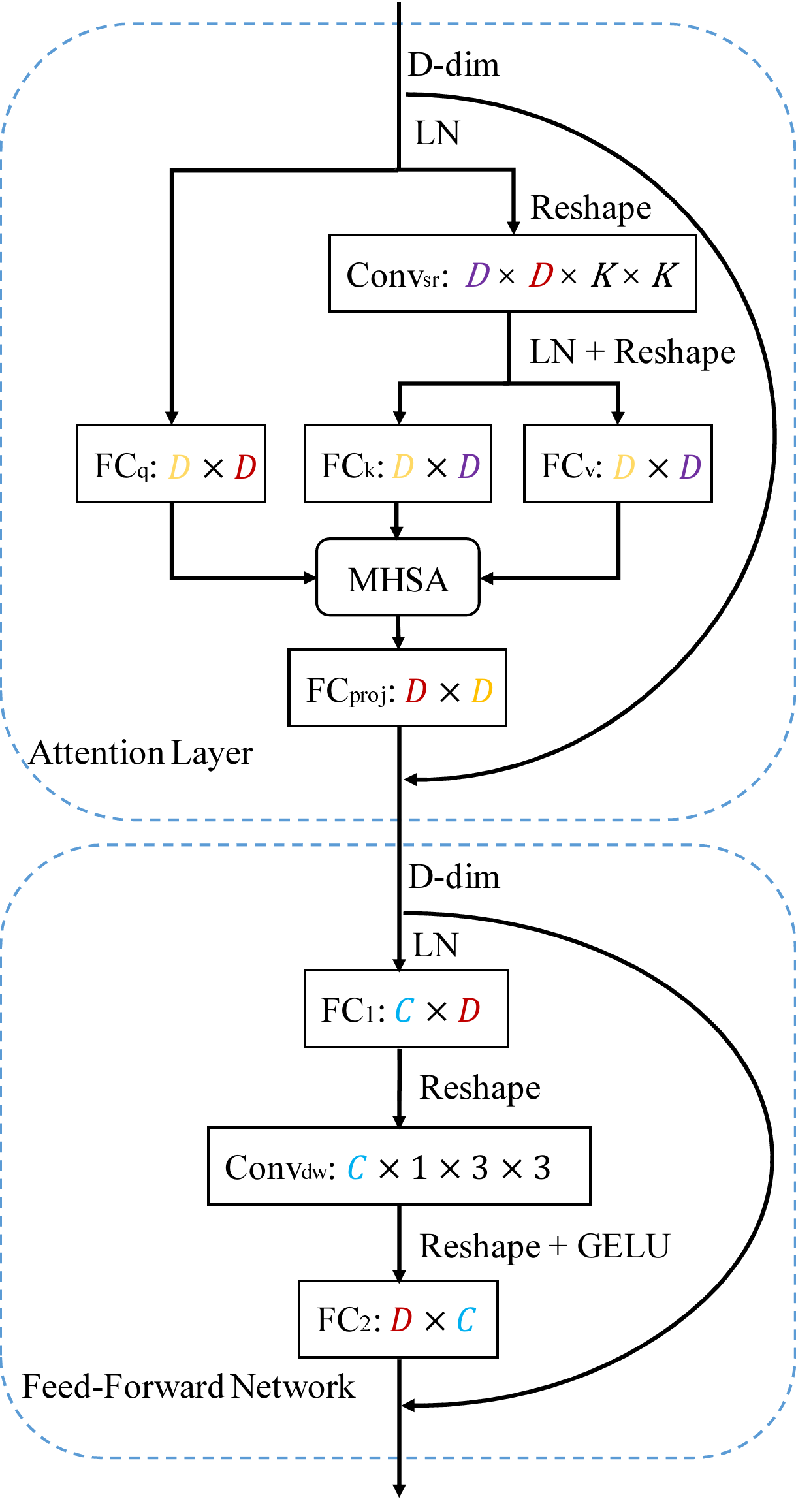}
  \caption{Illustration of four irrelevant components in a PVTv2 block. }
  \label{pipleine_pvtv2}
\end{figure}

\squishlist
    \item Component 1: The numbers shown in red, namely the shortcut connections that chain representations in every \emph{stage}, i.e., the input channels of $\mathrm{FC}_q$, $\mathrm{Conv}_{sr}$ and $\mathrm{FC}_1$, the output channels of $\mathrm{FC}_{proj}$ and $\mathrm{FC}_2$, and the first and third $\mathrm{LN}$ layers;
    \item Component 2: The numbers shown in gold, namely the attention embedding filters inside the attention layer in every \emph{block}, i.e., the input channels of $\mathrm{FC}_{proj}$ and the output channels of $\mathrm{FC}_q$, $\mathrm{FC}_k$ and $\mathrm{FC}_v$;
    \item Component 3: The numbers shown in blue, namely the FFN inter-layer filters in every \emph{block}, i.e., the input channels of $\mathrm{FC}_2$, the first channel of $\mathrm{Conv}_{dw}$ and the output channels of $\mathrm{FC}_1$.
    \item Component 4: The numbers shown in purple, namely the spatial reduction attention embedding filters inside the attention layer in every \emph{block}, i.e.,  the input channels of $\mathrm{FC}_k$ and $\mathrm{FC}_v$, the output channels of $\mathrm{Conv}_{sr}$, and the second $\mathrm{LN}$ layer.
\squishend

Note that because there are three additional patch embedding modules, when calculating the importance scores of component 1, we divide the whole shortcut connections into four stages and calculate importance separately. In particular, three are only three components in the final stage because the $\mathrm{Conv}_{sr}$ layer is removed.  
\section{Pruning PVTv2-B2 into PVTv2-B1}

When pruning PVTv2-B2 into PVTv2-B1, we compare two distillation methods. The first is the classic soft distillation and we set $\alpha$ in Equation~\ref{eq:KD} as 0.4. The second method is, as described in Section~\ref{Pruning_PVTv2}, we use the MSE loss to distill model. We show the results in Table~\ref{distill_pvt}. When distilling model with the MSE loss, UP-PVTv2-B1 obtains slightly higher accuracy than that using the KL loss.
\begin{table}[t]
  \centering
    \caption{Results of two distillation methods when pruning PVTv2-B2 into PVTv2-B1.  We fine-tuned UP-PVTv2-B1 model with 50 epochs. }
    \begin{tabular}{c|cc}
      \bottomrule[1pt]
    Methods    & Distillation & Accuracy  \\ \hline
    PVTv2-B1  &  - & 78.62\% \\ \hline
    \multicolumn{1}{c|}{\multirow{2}{*}{UP-PVTv2-B1}}  &  KL & 79.23\% \\ 
    \multicolumn{1}{c|}{}  & MSE & \textbf{79.48\%} \\ \hline
    \toprule[1pt]
    \end{tabular}
\label{distill_pvt}
\end{table}

\end{appendix}

\end{document}